%% file: neurips_2025.tex
\documentclass{article}
\PassOptionsToPackage{numbers, compress}{natbib}
\usepackage[dandb]{neurips_2025}

\usepackage[utf8]{inputenc} 
\usepackage[T1]{fontenc}    
\usepackage{hyperref}       
\usepackage{url}            
\usepackage{booktabs}       
\usepackage{amsfonts}       
\usepackage{nicefrac}       
\usepackage{microtype}      
\usepackage{xcolor}         
\usepackage{amsmath}
\usepackage{mathabx} 
\usepackage{graphicx}
\usepackage{tcolorbox}
\usepackage{colortbl}
\usepackage{subcaption}
\usepackage{float}
\usepackage{amssymb}
\usepackage{pifont}
\usepackage{multirow}
\usepackage[capitalise]{cleveref}

\DeclareMathOperator{\His}{\mathcal{H}}

\title{DanmakuTPPBench: A Multi-modal Benchmark for Temporal Point Process Modeling and Understanding}

\author{
  Yue Jiang$^1$, 
  Jichu Li$^2$, 
  Yang Liu$^3$, 
  Dingkang Yang$^1$, 
  Feng Zhou$^{2,4}$\thanks{Corresponding authors.}, 
  Quyu Kong$^3$\footnotemark[\value{footnote}]\\
  $^1$Fudan University\\
  $^2$Center for Applied Statistics and School of Statistics, Renmin University of China\\
  $^3$Independent Researcher\\
  $^4$Beijing Advanced Innovation Center for Future Blockchain and Privacy Computing\\
  \texttt{jiangyue23@m.fudan.edu.cn, feng.zhou@ruc.edu.cn} \\
  \texttt{kongquyu@gmail.com}
}

\begin{document}

\maketitle

\vspace{-10pt}

\begin{abstract}
We introduce \textit{DanmakuTPPBench}, a comprehensive benchmark designed to advance multi-modal Temporal Point Process (TPP) modeling in the era of Large Language Models (LLMs). 
While TPPs have been widely studied for modeling temporal event sequences, existing datasets are predominantly unimodal, hindering progress in models that require joint reasoning over temporal, textual, and visual information. To address this gap, \textit{DanmakuTPPBench} comprises two complementary components:
(1) \textit{DanmakuTPP-Events}, a novel dataset derived from the Bilibili video platform, where user-generated bullet comments (Danmaku) naturally form multi-modal events annotated with precise timestamps, rich textual content, and corresponding video frames;
(2) \textit{DanmakuTPP-QA}, a challenging question-answering dataset constructed via a novel multi-agent pipeline powered by state-of-the-art LLMs and multi-modal LLMs (MLLMs), targeting complex temporal-textual-visual reasoning. 
We conduct extensive evaluations using both classical TPP models and recent MLLMs, revealing significant performance gaps and limitations in current methods’ ability to model multi-modal event dynamics. Our benchmark establishes strong baselines and calls for further integration of TPP modeling into the multi-modal language modeling landscape. Project page: \url{https://github.com/FRENKIE-CHIANG/DanmakuTPPBench}. 
\end{abstract}

\section{Introduction}

Temporal Point Processes (TPPs) offer a powerful framework for modeling event sequences in continuous time and have shown effectiveness across a wide range of domains, including social media activity prediction, healthcare monitoring, earthquake modeling, and financial transaction analysis~\citep{Mishra2016FeaturePrediction,kong2019modeling,farajtabar2015coevolve,hawkes1971spectra,johnson1996point,bacry2015hawkes}. However, conventional TPP models often struggle to capture the rich multi-modal patterns and contextual dependencies present in real-world event streams. 
Meanwhile, Large Language Models (LLMs) and Multi-modal LLMs (MLLMs) have recently achieved impressive success across various tasks~\citep{openai2022,achiam2023gpt,jiang2024comt}. Integrating TPP modeling into these architectures presents a promising yet largely underexplored direction, with the potential to enhance temporal reasoning capabilities and unlock novel downstream applications~\citep{liu2024tpp,kong2025language, shi2023language}.

Existing TPP datasets~\citep{ni2019justifying,zhou2020fast,xue2023easytpp} were not designed with multi-modality in mind, resulting in a critical gap in benchmark resources for evaluating multi-modal TPP models. Most current datasets focus exclusively on temporal and categorical event attributes, overlooking the rich contextual information—such as text and visual signals—that frequently accompanies real-world event streams. This limitation constrains progress in developing models capable of joint reasoning across temporal, textual, and visual modalities. 
To address this gap, we propose \textit{DanmakuTPPBench}, a new benchmark comprising two datasets: \textit{DanmakuTPP-Events}, for conventional multi-modal TPP modeling, and \textit{DanmakuTPP-QA}, a novel question-answering dataset designed to assess deeper temporal and cross-modal understanding.

\begin{figure}[!tbp]
    \centering
    \includegraphics[width=\textwidth]{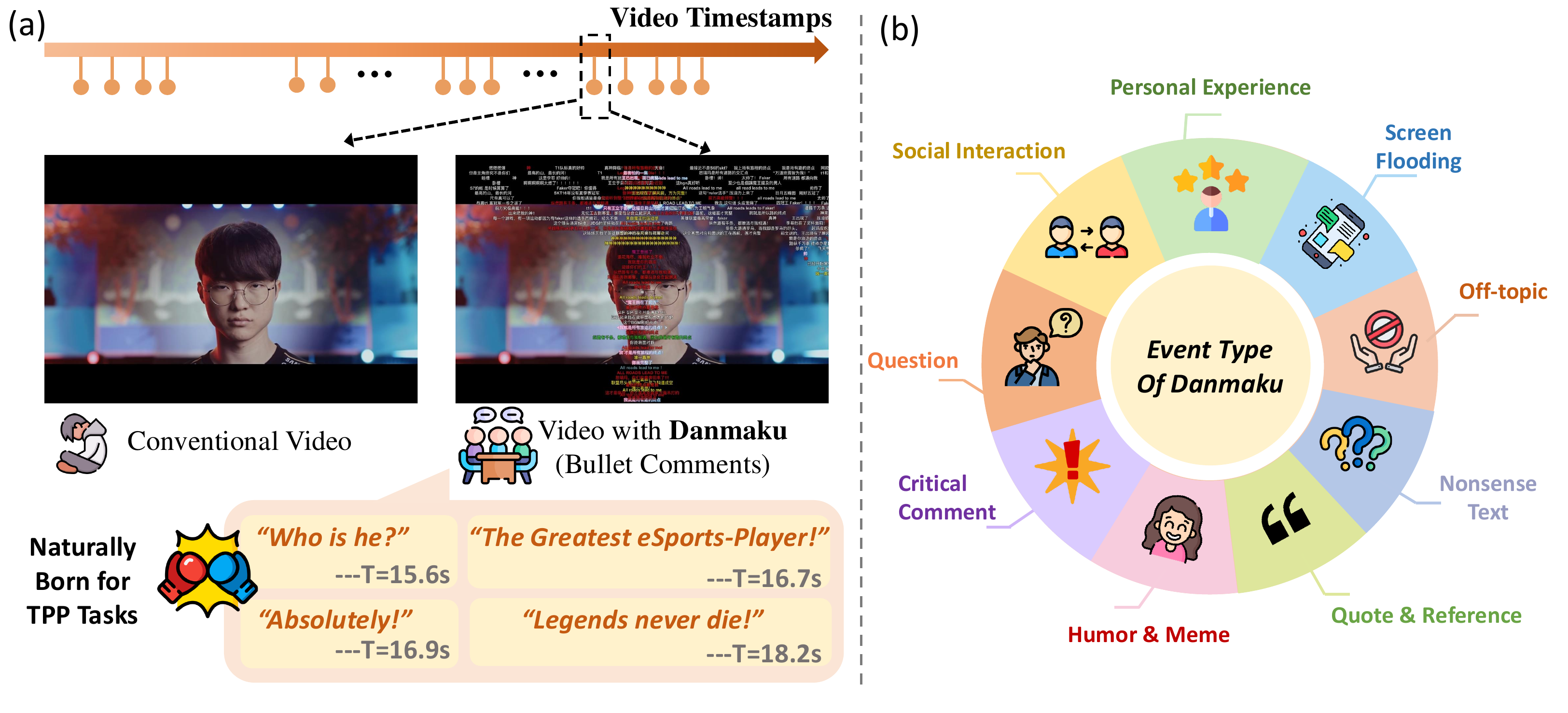}
    \caption{Introduction to Danmaku TPP data. (a) Comparison between conventional video viewing and Danmaku viewing experience. Danmaku appears as overlaid text messages at specific timestamps during video playback, creating a multi-modal TPP. The example shows comments from an esports video with timestamps. (b) Danmaku event types identified in our dataset.}
    \label{fig-teaser}
\end{figure}

We first introduce \textit{DanmakuTPP-Events}, a novel multi-modal TPP dataset constructed from Bilibili’s Danmaku (real-time bullet comment) system. Danmaku refers to user-generated comments that appear as overlaid text at specific timestamps during video playback, creating an interactive and communal viewing experience, as illustrated in Fig.~\ref{fig-teaser}~(a).
As one of China’s leading video-sharing platforms, Bilibili offers an ideal source for TPP data—each Danmaku comment naturally forms an event with precise temporal alignment, rich textual content, and visual context from video frames. This makes it a native setting for modeling temporal, textual, and visual modalities in combination. The resulting dataset contains 7,250 TPP sequences and over 10.8 million Danmaku events. 

While MLLMs excel at integrating diverse modalities and generating fluent textual responses, their ability to understand and reason about temporal point processes remains largely unexplored. To address this gap, we introduce \textit{DanmakuTPP-QA}, a question-answering dataset built on top of \textit{DanmakuTPP-Events}, specifically designed to evaluate temporal-visual-textual reasoning. 
We develop a novel multi-agent pipeline that leverages state-of-the-art LLMs and MLLMs—such as Deepseek-R1~\citep{guo2025deepseek-r1} for task generation and Qwen2.5-VL~\citep{yang2024qwen25} for visual understanding—to automatically construct 10 diverse evaluation tasks. The resulting dataset comprises a wide range of challenging open-ended and closed-ended questions, requiring fine-grained, multi-modal reasoning grounded in temporal dynamics.

Through extensive evaluation of both conventional TPP models and state-of-the-art MLLMs, we show that our benchmark poses significant challenges to existing approaches. The results highlight substantial room for improving how language models understand temporal point processes and perform integrated temporal reasoning across modalities.

With the introduction of \textit{DanmakuTPPBench}, our contributions are threefold:

(1) We present \textit{DanmakuTPP-Events}, the first multi-modal TPP dataset that jointly captures temporal, textual, and visual information from synchronized user comments and video content.

(2) We construct \textit{DanmakuTPP-QA}, a challenging question-answering benchmark generated via a novel LLM-powered multi-agent pipeline to assess temporal-visual-textual reasoning.

(3) We conduct comprehensive benchmark evaluations, uncovering key limitations in existing models and establishing strong baselines to guide future research in multi-modal TPP modeling.

\section{Related Works}

We summarize widely used datasets for TPP modeling in~\cref{tab-dataset}. These datasets primarily focus on capturing temporal patterns and event types across various application domains.
The \textbf{Retweet} dataset~\citep{zhou2013learning2} contains user retweet sequences, categorized into three groups based on follower counts. The \textbf{StackOverflow} dataset~\citep{jure2014snap} records user badge-awarding events with 22 distinct badge types. The \textbf{Taobao} dataset~\citep{xue2022byt5} captures user click behavior across 17 item categories, while the \textbf{Taxi} dataset~\citep{whong2014foiling} logs pick-up and drop-off events in New York City, categorized into 10 location-based event types. Although these datasets provide valuable temporal signals, they lack multi-modal context.
The \textbf{ActiveRT} dataset~\citep{Rizoiu2016ExpectingPopularity} contains tweets linking to YouTube videos collected over six months, introducing some content-level variation. The \textbf{RNCNIX}\citep{kong2020describing} and \textbf{Amazon Review}\citep{ni2019justifying} datasets incorporate textual information—news articles and customer reviews, respectively—offering partial multi-modality. However, these datasets still lack visual components, limiting their applicability for modeling richly contextualized event sequences.
In contrast, our proposed \textbf{\textit{DanmakuTPP-Events}} dataset is the first to natively integrate temporal, textual, and visual modalities by leveraging timestamped user comments overlaid on video content. This enables more comprehensive multi-modal TPP modeling in a naturally aligned setting.

\input{table/tab-datasets}

To the best of our knowledge, no existing benchmark provides a question-answering (QA) dataset specifically designed for TPPs. However, recent efforts in time series analysis—a closely related domain—have begun to explore QA-based evaluation. For example, \citet{kong2025time} proposed \textbf{TSQA}, a large-scale dataset comprising approximately 200k QA pairs across diverse time series domains such as weather and traffic. Similarly, \textbf{ECG-QA}~\citep{oh2023ecg} targets electrocardiogram interpretation through expert-validated QA pairs covering clinically relevant topics. 
Building on these ideas, we introduce \textbf{\textit{DanmakuTPP-QA}}, a novel QA benchmark that goes beyond conventional TPP modeling to assess temporal-visual-textual reasoning capabilities. It comprises 10 diverse task types designed to challenge and evaluate the ability of language models to integrate information across multiple modalities—addressing a significant gap in current TPP benchmarks. 

\section{DanmakuTPPBench}
To bridge the gap in multi-modal TPP datasets and address the limited progress in TPP understanding, we introduce \textit{DanmakuTPPBench}—the first comprehensive benchmark supporting both multi-modal TPP modeling and multi-task TPP understanding. \textit{DanmakuTPPBench} comprises two complementary components:
(1) \textit{DanmakuTPP-Events}, a dataset for conventional TPP modeling; and
(2) \textit{DanmakuTPP-QA}, a curated benchmark designed to evaluate TPP understanding via a suite of QA tasks. 
In this section, we describe the construction pipeline and design principles behind \textit{DanmakuTPPBench} in detail. 

\begin{figure}[!tbp]
    \centering
    \includegraphics[width=\textwidth]{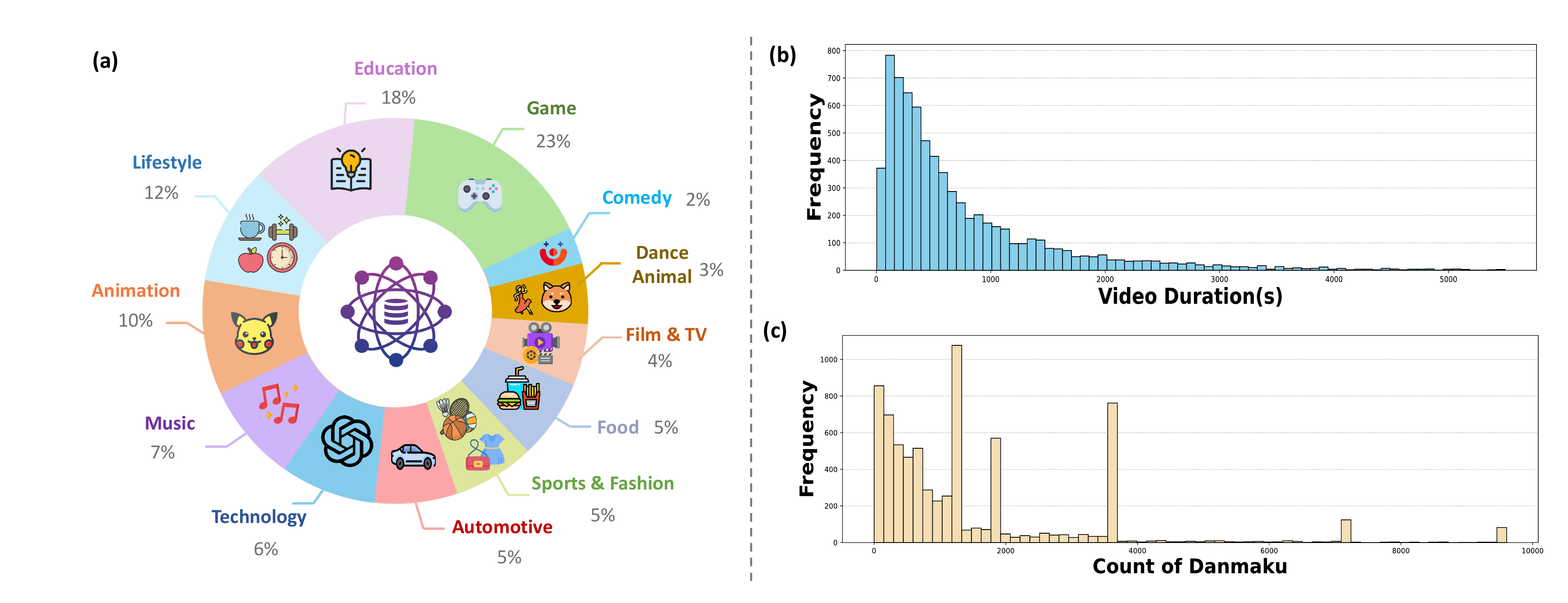}
    \caption{Statistics of  \textit{DanmakuTPP-Events} dataset. (a) The proportion of TPP data for video topics. (b) Distribution of video durations. (c) Distribution of Danmaku event count.}
    \label{fig-dataset}
\end{figure}

\subsection{Construction of DanmakuTPP-Events Dataset}

\textbf{Data Collection.}
We select Danmaku data as the foundation for our benchmark, as it inherently represents a multi-modal temporal point process. The Danmaku system enables users to post live comments that are overlaid on video playback at specific timestamps, naturally forming temporal events enriched with textual content and corresponding video frames. This seamless integration of temporal, textual, and visual modalities provides a unique opportunity to study complex interactions between user engagement patterns and rich contextual signals—an aspect that traditional TPP datasets fail to capture.

Bilibili is a leading Chinese video-sharing platform where users can upload, watch, and annotate videos with Danmaku comments. We manually collect all videos posted by the top 100 most popular creators on Bilibili in 2024\footnote{The full list of creators is available at \url{https://www.bilibili.com/BPU2024}.}, resulting in a total of 7,250 videos. 
From these videos, we construct a large-scale multi-modal TPP dataset, \textit{DanmakuTPP-Events}, which includes 10,820,790 Danmaku events. Formally, for a given video $v$, the Danmaku sequence is represented as $\His_v = {(t_i, e_i, m^{\text{text}}_i, m^{\text{image}}_i)}_{i=1}^N$, where each event $i$ comprises a timestamp $t_i$, event type $e_i$, a textual mark $m^{\text{text}}_i$, and an associated video frame $m^{\text{image}}_i$. We preprocess and format the dataset following the protocol established in~\citep{xue2023easytpp}.

\textbf{Ethical Issues.} During the data collection process, we strictly adhered to Bilibili’s Terms of Service and only collected data from publicly available videos that did not require any login or authentication.  Furthermore, regarding data privacy and user anonymization, all Danmaku comments in the interactive videos are entirely dissociated from any identifiable user information. The dataset we constructed contains only Danmaku text, temporal sequences, and video frames, and excludes any user account details, personally identifiable information (PII), or other privacy-sensitive content. Finally, to mitigate potential harassment or harmful content that may exist in raw data from the video platform, we proactively implemented multiple methodological precautions to ensure ethical compliance, as detailed in the Appendix~\ref{appendix-ethical}.

\textbf{Data Statistics.}
As shown in \cref{fig-dataset} (a), \textit{DanmakuTPP-Events} spans fourteen distinct video categories. Gaming constitutes the largest share (23\%), followed by education (18\%), lifestyle (12\%), and animation (10\%). This diverse topical distribution ensures broad domain coverage, enhancing the dataset’s ecological validity for TPP modeling.

\cref{fig-dataset} (b) presents the distribution of video durations, revealing that most videos fall within the 0–500 second range. The distribution exhibits a long-tail pattern, with a small number of videos exceeding 1000 seconds and a few reaching up to 5000 seconds. These variations introduce significant diversity in the temporal structure of Danmaku activity.

\cref{fig-dataset} (c) shows the distribution of Danmaku comment counts per video. While the majority of videos receive 500–2000 comments, there are pronounced peaks around 1000 and 3500, with some videos attracting 8000–10000 comments. This wide variation in comment density gives rise to a rich variety of temporal patterns, ranging from dense comment bursts to more dispersed interactions over time.

Together, the heterogeneity in video topics, durations, and comment densities contributes to the complexity and expressiveness of the \textit{DanmakuTPP-Events} dataset, offering not just breadth of content, but depth of interaction, and making it a valuable resource for advancing multi-modal TPP modeling. 

\subsection{Multi-Agent Collaboration for DanmakuTPP-QA Construction}

\begin{figure}[!tbp]
    \centering
    \includegraphics[width=\textwidth]{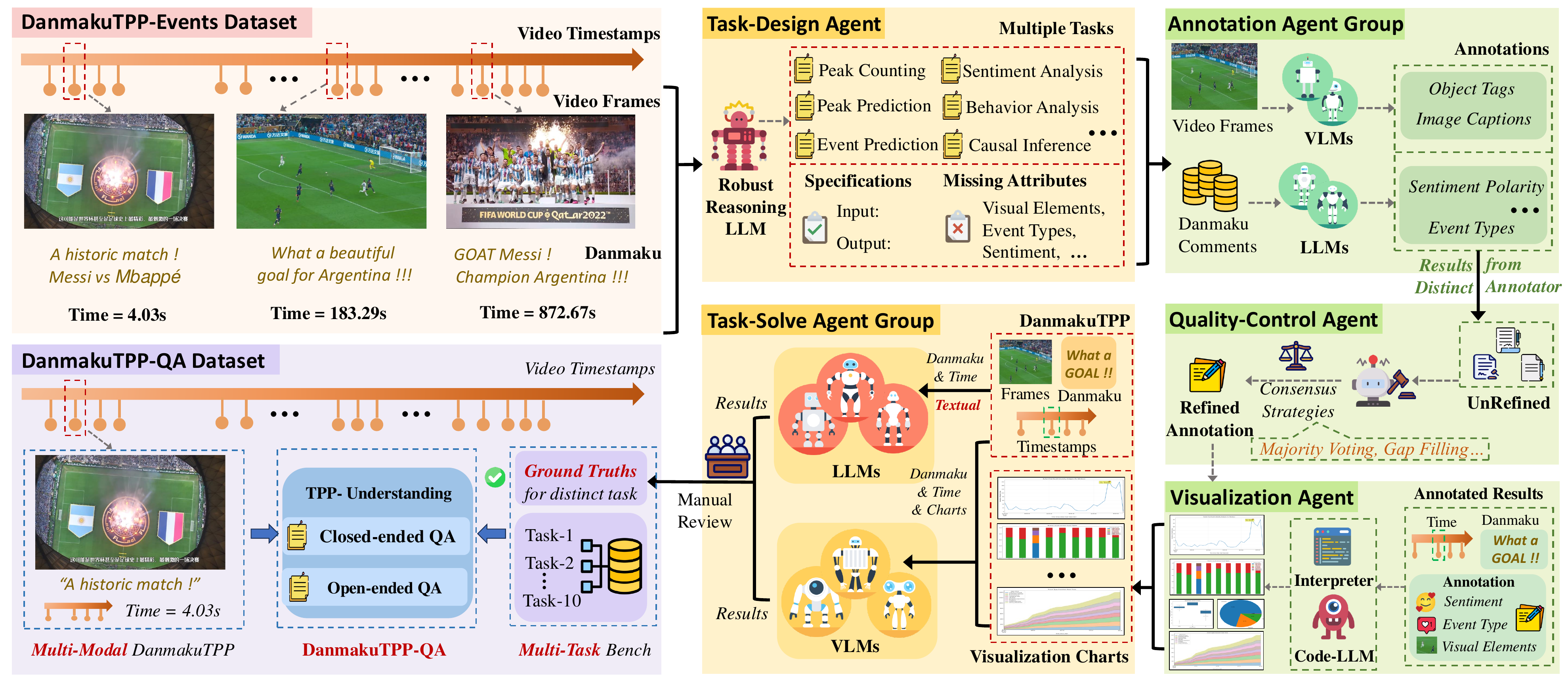}
    \caption{Multi-agent framework for automated construction of DanmakuTPP-QA. The framework consists of five main components: (1) \textit{DanmakuTPP-Events} (top left) containing synchronized video frames, timestamps, and user comments; (2) Task-Design Agent employing a reasoning LLM to generate diverse evaluation tasks; (3) Annotation Agent Group extracting object tags, image captions, sentiment polarity, and event types; (4) Quality-Control Agent implementing consensus strategies to refine annotations through majority voting and gap filling; (5) Task-Solve Agent Group solving the designed tasks based on multi-modal inputs. This framework enables the creation of \textit{DanmakuTPP-QA} covering multiple tasks with ground truths.}
    \label{fig-framework}
\end{figure}

To enable temporal-textual-visual reasoning, we curated a subset of 2,605 videos from the \textit{DanmakuTPP-Events} dataset, each containing between 500 and 1,500 Danmaku events. Based on this selection, we developed a multi-agent collaborative framework for automated data construction, resulting in a challenging multi-task benchmark: \textit{DanmakuTPP-QA}.
As shown in~\cref{fig-framework}, our pipeline integrates five specialized agents, each responsible for a key stage in the data generation process: 

\textbf{(1) Task-design Agent.}
This agent formulates meaningful research tasks by analyzing the structure and attributes of Danmaku-based TPP data. We use Deepseek-R1~\cite{guo2025deepseek-r1} for its strong reasoning ability, prompting it to act as domain experts (e.g., social scientists or network analysts). It defines task goals, input-output formats, and identifies attribute limitations in the dataset. Applied to \textit{DanmakuTPP-Events}, this agent designed 10 distinct tasks, including 8 closed-ended and 2 open-ended formats.

\textbf{(2) Annotation Agent.}
Responsible for labeling each TPP event based on task definitions, this module uses both textual and visual models. Text annotations are generated using Qwen2.5~\cite{yang2024qwen25}, while visual annotations are produced using Qwen2.5-VL~\cite{Qwen2.5-VL} and the Recognize Anything Model (RAM)~\cite{zhang2023recognizeanythingstrongimage}. This ensemble strategy ensures high-quality, multi-modal annotations across Danmaku comments and corresponding video frames.

\textbf{(3) Quality-control Agent.}
This agent validates annotation consistency by aggregating and reconciling outputs from multiple Annotation Agents. We utilize Qwen3~\cite{yang2025qwen3technicalreport} to compare results, apply filtering, and consolidate labels. When conflicts arise, majority voting and rule-based gap-filling strategies are employed. The final outputs are integrated with the original \textit{DanmakuTPP-Events} data to form the curated QA dataset.

\textbf{(4) Visualization Agent.}
It transforms temporal sequences and annotated attributes into interpretable visualizations. Using Qwen2.5-Coder~\cite{hui2024qwen25coder}, Python scripts are automatically generated to produce charts such as Danmaku intensity timelines, sentiment evolution plots, and event-type distributions. These visualizations enrich the raw data with aligned textual-temporal-visual representations.

\textbf{(5) Task-solve Agent.}
To generate reference answers, this agent group analyzes both the raw TPP data and the visualizations. We incorporate multiple advanced LLMs and MLLMs, including Qwen3~\cite{yang2025qwen3technicalreport}, Qwen2.5-VL~\citep{yang2024qwen25}, and Gemma-3~\citep{team2025gemma3}. LLMs extract temporal dynamics from textual sequences, while MLLMs align event information with visualizations, leveraging their capabilities in chart interpretation and multi-modal reasoning. Final answers are obtained via majority voting, supplemented by gap-filling and manual verification, and are used as ground-truth labels in \textit{DanmakuTPP-QA}.

The following is a sample output from Task-solve Agent addressing a question provided by Task-design Agent:

\begin{tcolorbox}[
  colback=lightgray!30,
  colframe=black,
  arc=0mm,
  boxrule=0.5pt,
  left=5pt,
  right=5pt,
  top=5pt,
  bottom=5pt,
  width=\textwidth,
]
\textbf{Question:}\\
$<|TPP\_data|>$

Please provide a detailed analysis of the sentiment trend in the Danmaku comments over time. 

\textbf{Answer:}\\
The sentiment trend starts negative but quickly rises to a stable positive polarity around 0.4, with minor fluctuations. Significant inflection points include an initial sharp rise at 01:00:01 and a peak near 01:00:04. The trend remains consistently positive thereafter, indicating a generally positive viewer reaction. Abrupt changes are minimal, suggesting a steady emotional response throughout the video. Major sentiment shifts occur only at the beginning, transitioning from negative to positive.
\end{tcolorbox}
where $<|TPP\_data|>$ serves as a placeholder for the corresponding TPP data associated with each video. Detailed prompt designs for all agents are provided in Appendix~\ref{appendix-a}. 
To ensure evaluation accuracy, all answers generated by the Task-solve Agent group for the test set underwent rigorous manual verification. This process guarantees the high fidelity of the ground-truth annotations, establishing a reliable foundation for downstream model evaluation and performance benchmarking.

\textbf{Composition.}
\textit{DanmakuTPP-QA} is divided into three subsets: a training set (2,005 samples), a validation set (300 samples), and a test set (300 samples). As summarized in~\cref{tab-task}, the benchmark comprises 10 distinct tasks, including 8 TPP-focused QA tasks and 2 higher-level temporal-textual-visual reasoning tasks. These tasks collectively span a wide range of challenges, such as predicting the timing and burst peaks of Danmaku events, classifying their trigger types, analyzing sentiment trends, and identifying the underlying causes of Danmaku bursts. 
\input{table/tab-task}

\section{Experiments}

In this section, we present a comprehensive evaluation of state-of-the-art models on our proposed \textit{DanmakuTPPBench}, demonstrating both the challenges and opportunities in modeling multi-modal temporal point processes.

\subsection{Experimental Settings}
\textbf{DanmakuTPP-Events Evaluation.} We employ the following state-of-the-art deep TPP models when evaluating TPP tasks on DanmakuTPP: (1) Neural Hawkes Process (NHP)~\citep{mei2017neural} and Recurrent Marked Temporal Point Process (RMTPP)~\citep{du2016RMTPP} are two RNN-based models, encoding the temporal and type information of historical events. As for NHP, the resulting history embedding is used to model the conditional intensity function. 
(2) Self-Attentive Hawkes Process (SAHP)~\citep{zhang2020self}, Transformer Hawkes Process (THP)~\citep{zuo2020transformer} and Attentive neural Hawkes proces (AttNHP)~\citep{attnhp-yang2022transformerembeddingsirregularlyspaced}:
all three models encode historical events through self-attention mechanism.
We use the implementation provided by EasyTPP~\citep{xue2023easytpp}.  
(3) ODE-based TPP (ODETPP)~\citep{chen2020ODETPP} models the continuous-time evolution of a latent state via a neural ordinary differential equation. It can be viewed as a spatio-temporal point process restricted to the temporal dimension, with the spatial component omitted.
All models are trained with the default setting in the EasyTPP framework to ensure a fair and standardized comparison. We note that the multi-modal information is not used for these models.

\textbf{DanmakuTPP-QA Evaluation.} We frist evaluate the TPP understanding capabilities of existing open source pretrained LLMs (Qwen2.5 series~\citep{yang2024qwen25}, Qwen3 series~\citep{yang2025qwen3technicalreport} and Deepseek-V3~\citep{liu2024deepseek-v3}) and MLLMs (Gemma3~\citep{team2025gemma3} and  Qwen2.5-VL~\citep{wang2024qwen2vl}). According to the different video clips required by each task, we provide the textual timestamps and Danmaku content within the time window as inputs to both LLMs and MLLMs. Inspired by the mainstream paradigm for video understanding employed by several powerful MLLMs, we process a video as a sequence input of discrete image frames.  We randomly sample three video frames within the time window for the supplementary input of MLLMs, enhancing the multi-modal understanding. The strategy of using three frames was intended as an efficient starting point, balancing the need for visual context against the significant computational and memory demands associated with processing long TPP sequences combined with numerous video frames. Consequently, LLMs focus on mining temporal evolution patterns from timestamps and Danmaku, while MLLMs are responsible for aligning textual TPP information with video frames to address complex TPP tasks.

We then further evaluate the model performance after finetuning. We conduct LoRA~\citep{hu2022lora} finetuning with Qwen2.5-VL-3B. For each task, we train the model on a single NVIDIA RTX 4090 for 3 epochs with learning rate 1e-4. Due to GPU memory constraints, we set a truncation of the sequence for some tasks during training. Details about hyperparameters can be found in Appendix~\ref{appendix-b}.

\textbf{Evaluation Metrics.} For conventional TPP tasks on  \textit{DanmakuTPP-Events}, we employ RMSE for evaluating next-event timestamp prediction and test log-likelihood values for measuring modeling performance \citep{xue2023easytpp}. For \textit{DanmakuTPP-QA}, we use different metrics to comprehensively assess model performance across various task settings, as shown in~\cref{tab-task}. Tasks related to temporal prediction and sentiment polarity computation adopt root mean square error (RMSE). Tasks related to Danmaku event type prediction are evaluated by accuracy. Open-ended tasks are evaluated using Qwen3-235B-A22 to output a correctness score between 0-1 when compared against ground-truth descriptions.

\subsection{Results on DanmakuTPP-Events}

\begin{figure}[!tbp]
    \centering
    \begin{subfigure}[b]{0.48\textwidth}
        \centering
        \includegraphics[width=\textwidth]{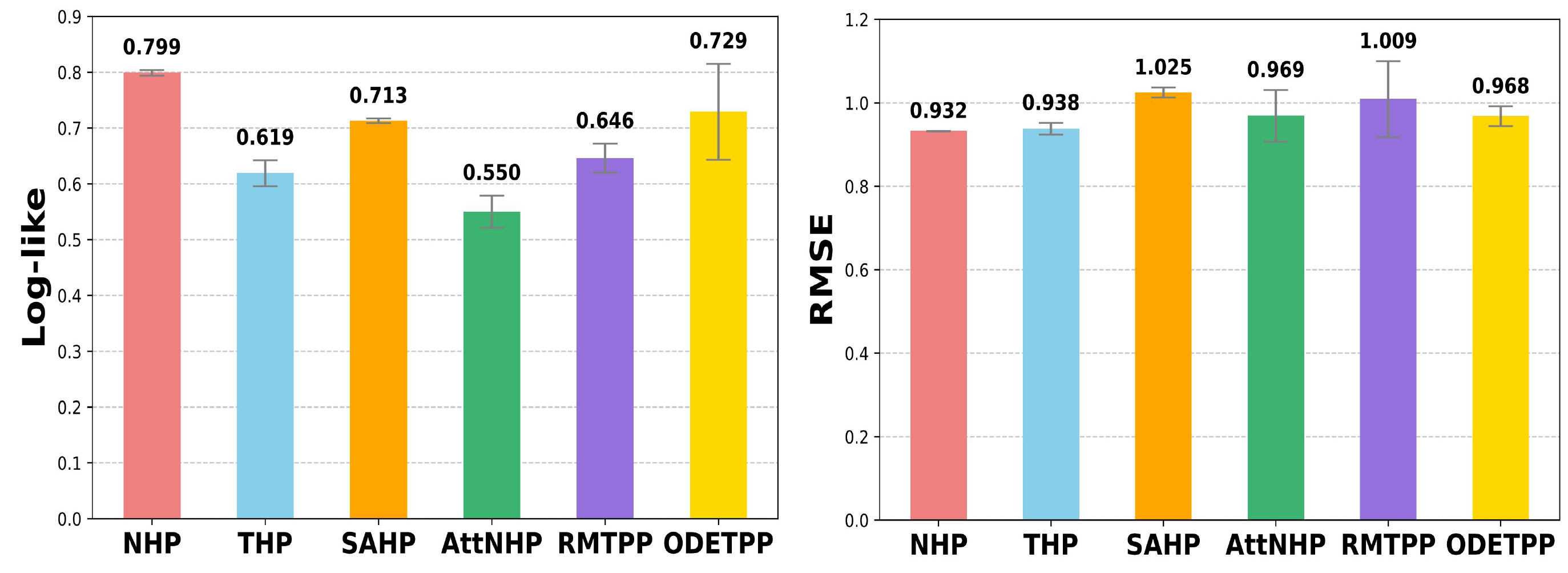}
        \caption{}
        \label{fig-easytpp}
    \end{subfigure}
    \hfill
    \begin{subfigure}[b]{0.48\textwidth}
        \centering
        \includegraphics[width=\textwidth]{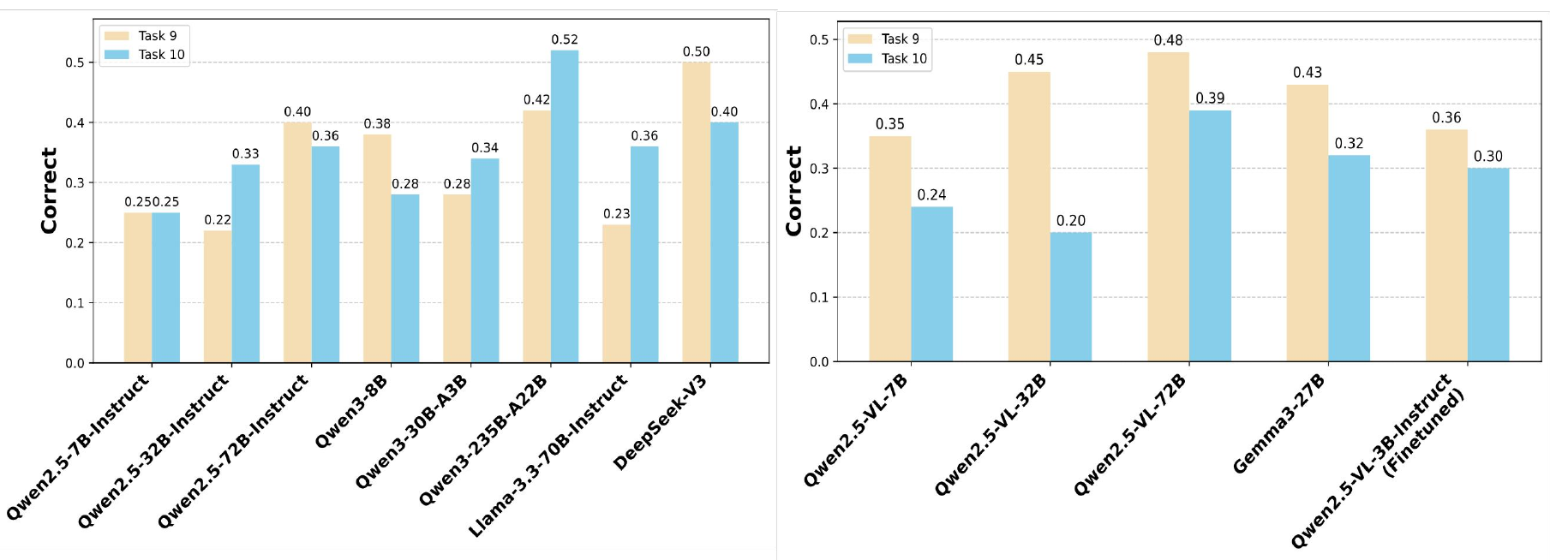}
        \caption{}
        \label{fig-open-result}
    \end{subfigure}
    \caption{Evaluations of TPP models: (a) Conventional TPP Models on the \textit{DanmakuTPP-Events} dataset; (b) LLM-based Evaluation of LLMs and MLLMs on DanmakuTPP-QA open-ended TPP Questions. The correctness of answers is scored from 0 to 1 by Qwen3-235B-A22B~\cite{hui2024qwen25coder}.}
    \label{fig-combined}
\end{figure}

As shown in~\cref{fig-easytpp}, we observe varying performance across different TPP models on \textit{DanmakuTPP-Events} dataset. For log-likelihood evaluation, NHP achieves the best performance with a score of 0.799, followed by ODETPP (0.729), SAHP (0.713), and RMTPP (0.646), while THP (0.619) and AttNHP (0.550) show comparatively lower performance. This suggests that the continuous-time LSTM approach in NHP may be more effective at capturing the complex temporal dynamics in Danmaku data than some attention-based approaches.

For next-event prediction evaluated by RMSE where the lower is better, NHP and THP perform best with scores of 0.932 and 0.938, respectively, followed closely by ODETPP (0.968) and AttNHP (0.969). SAHP and RMTPP show the highest error with RMSE scores of 1.025 and 1.009, respectively. Taken together, while NHP performs best on both metrics, discrepancies for models like SAHP and THP suggest that modeling the event-time distribution and optimizing next-event accuracy emphasize different inductive biases, making dual-metric evaluation essential.

\subsection{Results on DanmakuTPP-QA}
\Cref{tab-exp-closed} presents a comprehensive evaluation of various LLMs and MLLMs on the \textit{DanmakuTPP-QA} dataset across eight closed-ended TPP tasks. The tasks range from temporal prediction (T-1 to T-3) to sentiment analysis (T-4 to T-6) and event type prediction (T-7 to T-8), collectively assessing models' capabilities in understanding and reasoning about multimodal temporal point processes. The results reveal several important insights about model performance on TPP understanding tasks.

\textbf{Model Scaling Effects.} We observe clear benefits from scaling model size within the same model family. For instance, in the Qwen2.5-Instruct series, the 32B and 72B variants consistently outperform the 7B model across most tasks. This is particularly evident in Task-2 (next Danmaku timestamp prediction), where RMSE decreases dramatically from 27.64 (7B) to 1.52 (32B) and 1.28 (72B). Similarly, in the Qwen2.5-VL series, performance generally improves with model size, with the 72B variant achieving the best results in Tasks 2, 3, 7, and 8. These suggest that larger models capacity helps capture the complex temporal dynamics and multi-modal relationships in Danmaku data. This scaling pattern aligns with findings in other domains where increased parameter counts improve performance on tasks requiring complex reasoning and pattern recognition.

\input{table/tab-exp-closed}

\textbf{LLMs v.s. MLLMs.} We do not observe a consistent advantage for MLLMs over LLMs despite the multi-modal nature of the dataset. While MLLMs like Qwen2.5-VL-72B excel in certain tasks (achieving the highest accuracy of 47.17\% on Task-8), LLMs often perform competitively or better on others. For example, Llama-3.3-70B-Instruct achieves the lowest RMSE (1.11) on Task-2, and Qwen3-30B-A3B achieves the highest accuracy (23.00\%) on Task-7. We believe that the current integration of visual information in MLLMs may not be optimal given long context lengths. This unexpected finding suggests that text-only models may effectively leverage linguistic cues to infer temporal patterns, and that current visual-temporal integration mechanisms in MLLMs require further refinement to fully utilize multimodal information.

\textbf{Model Family Comparison.} Among the model families, Qwen3 models demonstrate strong performance on sentiment-related tasks (Tasks 4-6), with Qwen3-30B-A3B achieving the best RMSE (0.20) on Task-4. DeepSeek-V3 and Llama-3.3 excel in predicting sentiment polarity for future events (Tasks 5-6). For event type prediction (Tasks 7-8), Qwen3-30B-A3B and Qwen2.5-VL-72B achieve the highest accuracies, suggesting their superior ability to understand the relationship between temporal patterns and event categories. These performance differences between model families likely stem from variations in pretraining objectives, architectural design choices, and data composition, highlighting the importance of model selection for specific TPP tasks in real-world applications.

\textbf{Finetuning.} The finetuned Qwen2.5-VL-3B model, despite its relatively small size (3B parameters), outperforms all larger pretrained models on sentiment-related prediction tasks (Tasks 4-6), achieving RMSE of 0.05, 0.16, and 0.08, respectively. This represents a reduction in error by factors of 4-6× compared to the best pretrained models. The finetuned model also shows strong performance on Task-1 (27.0\% accuracy) and Task-8 (43.0\% accuracy). These results highlight the importance of task-specific adaptation for TPP understanding, suggesting that even smaller models can achieve superior performance when finetuned on relevant data.
However, we also observe that finetuning can lead to performance degradation in some cases, as seen in Task-3 where the finetuned model's RMSE (220.43) is significantly worse than all pretrained models. This suggests potential overfitting or optimization challenges when finetuning for certain temporal prediction tasks.

\textbf{Open-ended TPP Question Performance.} Beyond closed-ended tasks, we further evaluate models on open-ended TPP questions from Tasks 9 and 10, which require deeper reasoning about temporal dynamics. As shown in \cref{fig-open-result}, model performance varies significantly across these challenging tasks. For Task-9 (analysis of global sentiment dynamics), Qwen2.5-VL-72B and Qwen3-235B-A22B demonstrate the strongest performance with correctness scores of 0.48 and 0.42, respectively. In Task-10 (causal attribution analysis), Qwen3-235B-A22B achieves the highest score (0.52), substantially outperforming other models. Notably, we observe that while larger models generally perform better, model architecture and pretraining strategies also play crucial roles, as evidenced by DeepSeek-V3's competitive performance (0.40 on Task-10) despite its smaller parameter count compared to some other models.
The finetuned Qwen2.5-VL-3B-Instruct model (finetuned specifically on DanmakuTPP-QA) demonstrates decent performance on both open-ended tasks (0.36 and 0.30 on Tasks 9 and 10, respectively), outperforming several larger pretrained models. This further confirms that targeted finetuning on TPP data can effectively enhance models' ability to reason about complex temporal patterns, even with relatively limited parameter counts.

Overall, these results demonstrate that while current LLMs and MLLMs show promising capabilities in TPP understanding, there remains substantial room for improvement, particularly in tasks requiring precise temporal prediction and complex event sequence understanding. The strong performance of finetuned models on specific tasks suggests that targeted adaptation strategies may be crucial for advancing the state-of-the-art in multi-modal TPP modeling and understanding. The performance gap between different models on open-ended tasks further highlights the challenge of developing multi-modal LLMs for temporal-visual-textual reasoning with TPPs.


\section{Conclusions}
In this paper, we introduce \textit{DanmakuTPPBench}, the first comprehensive benchmark for multi-modal temporal point process modeling and understanding. Our benchmark addresses a significant gap in existing TPP research by incorporating rich multi-modal context—temporal, textual, and visual information—derived from Bilibili's Danmaku system. The benchmark consists of two complementary datasets:  \textit{DanmakuTPP-Events} for conventional TPP modeling and \textit{DanmakuTPP-QA}  for evaluating temporal-visual-textual reasoning capabilities across 10 diverse tasks.

Through our multi-agent collaborative workflow, we successfully construct a high-quality dataset that enables the systematic evaluation of how models reason about temporal patterns in conjunction with textual and visual modalities. Our extensive experiments with both conventional TPP models and state-of-the-art language models revealed significant challenges in multi-modal temporal reasoning, highlighting opportunities for future research. The performance gap between specialized TPP models and general-purpose language models demonstrates the need for improved integration of temporal point process understanding into multi-modal systems.

\textbf{Limitations:} Despite these contributions, our work has several limitations. First, the dataset is primarily sourced from Chinese-language content, which may limit its generalizability to other linguistic and cultural contexts. Second, while our multi-agent annotation pipeline significantly reduces manual effort, the quality of annotations remains dependent on the capabilities of the underlying models, which may introduce subtle biases or inconsistencies in the labeled data. Nevertheless, \textit{DanmakuTPPBench} represents a crucial step forward in multi-modal TPP research, providing opportunities to study the interplay between temporal dynamics and rich contextual information.

\section*{Acknowledgments}
This work was supported by the NSFC Project (No.62576346), the MOE Project of Key Research Institute of Humanities and Social Sciences (22JJD110001), the fundamental research funds for the central universities, and the research funds of Renmin University of China (24XNKJ13), and Beijing Advanced Innovation Center for Future Blockchain and Privacy Computing.

\bibliographystyle{plainnat}
\bibliography{ref}

\clearpage
\input{checklist}

\newpage
\appendix
\input{Appendix}

\end{document}

%% file: table/tab-datasets.tex
\begin{table}[t]
    \caption{Comparison between prior TPP datasets and our proposed \textit{DanmakuTPPBench}. \textit{K} denotes the number of event types, \textit{$L_{avg}$} stands for the average TPP sequence length. The Danmaku event types identified in our dataset are illustrated in Fig.~\ref{fig-teaser}~(b)}
    \begin{tabular}{lcccccc}
    \toprule
    \textbf{Dataset} & \textbf{Seq. No.} &  \textbf{\textit{K}} & \textbf{\textit{L$_{avg}$}}  & \textbf{Text Mark} & \textbf{Image Mark} & \multirow{1}{*}{\textbf{QA}} \\
    \midrule
    Retweet~\citep{zhou2013learning2} & 12,055 & 3 & 70 & \textcolor{red}{\ding{55}} & \textcolor{red}{\ding{55}} & \textcolor{red}{\ding{55}} \\
    Stackoverflow~\citep{jure2014snap} & 2,200 & 22 & 65 & \textcolor{red}{\ding{55}} & \textcolor{red}{\ding{55}} & \textcolor{red}{\ding{55}} \\
    Taobao~\citep{xue2022byt5} & 2,000 & 17 & 150 & \textcolor{red}{\ding{55}} & \textcolor{red}{\ding{55}} & \textcolor{red}{\ding{55}} \\
    Taxi~\citep{whong2014foiling} & 2,000 & 10 & 37 & \textcolor{red}{\ding{55}} & \textcolor{red}{\ding{55}} & \textcolor{red}{\ding{55}} \\
    ActiveRT~\citep{Rizoiu2016ExpectingPopularity} & 39,970 & - & 197 & \textcolor{red}{\ding{55}} & \textcolor{red}{\ding{55}} & \textcolor{red}{\ding{55}} \\
    RNCNIX~\citep{kong2020describing} & 8,129,126 & 2 & 7 & \textcolor{green}{\checkmark} & \textcolor{red}{\ding{55}} & \textcolor{red}{\ding{55}} \\
    Amazon Review~\citep{ni2019justifying} & 6,019 & 24 & 27 & \textcolor{green}{\checkmark} & \textcolor{red}{\ding{55}} & \textcolor{red}{\ding{55}} \\
    \hline
    \textit{DanmakuTPP-Events} & 7,250 & 9 & 1494 & \textcolor{green}{\checkmark} & \textcolor{green}{\checkmark} & \textcolor{red}{\ding{55}} \\
    \textit{DanmakuTPP-QA}  & 2,605 & 9 & 967 & \textcolor{green}{\checkmark} & \textcolor{green}{\checkmark} & \textcolor{green}{\checkmark} (10 tasks) \\
    \bottomrule
    \end{tabular}
    \label{tab-dataset}
    \end{table}

%% file: table/tab-task.tex
\begin{table}[t]
\caption{Supported tasks and corresponding evaluation metrics for the DanmakuClosed-ended multi-task dataset. For closed-ended tasks, we adopt accuracy and RMSE to evaluate the model's performance. For open-ended tasks, we employ a LLM for evaluation.}
\centering
\resizebox{\textwidth}{!}{
\begin{tabular}{clcc}
\toprule
 & \textbf{Task Description} & \textbf{Evaluation Metrics} & \textbf{Task Type} \\
\midrule
Task-1                 & \begin{tabular}[c]{@{}l@{}}Danmaku burst peak counting\end{tabular}                   & ACC                    & Closed-ended                    \\
Task-2                 & \begin{tabular}[c]{@{}l@{}}Prediction of the next\\ Danmaku timestamp\end{tabular}                                   &  RMSE                          & Closed-ended                    \\
Task-3                 & \begin{tabular}[c]{@{}l@{}}Prediction of the next Danmaku\\burst peak timestamp\end{tabular}                        &  RMSE                          & Closed-ended                    \\
Task-4                 & \begin{tabular}[c]{@{}l@{}}Assessment of average \\ sentiment polarity\end{tabular}                                      &  RMSE                    & Closed-ended                    \\
Task-5                 & \begin{tabular}[c]{@{}l@{}}Sentiment polarity prediction \\ for the next Danmaku\end{tabular}                         &   RMSE                    & Closed-ended                    \\
Task-6                 & \begin{tabular}[c]{@{}l@{}}Sentiment polarity prediction\\ for the next Danmaku burst peak\end{tabular}              &   RMSE                    & Closed-ended                    \\
Task-7                 & \begin{tabular}[c]{@{}l@{}}Event type inference for\\ the next Danmaku\end{tabular}                                   & ACC                    & Closed-ended                    \\
Task-8                 & \begin{tabular}[c]{@{}l@{}}Prediction of Top-2 triggering event\\ types for the next burst peak\end{tabular} & ACC                   & Closed-ended                    \\
\rowcolor[gray]{0.8} Task-9                 & \begin{tabular}[c]{@{}l@{}}Analysis of global sentiment\\ dynamics and the underlying drivers\end{tabular}            & LLM-Eval   & Open-ended         \\
\rowcolor[gray]{0.8} Task-10                & \begin{tabular}[c]{@{}l@{}}Causal attribution analysis for\\ specific Danmaku burst peak formation\end{tabular}         & LLM-Eval   & Open-ended         \\
\bottomrule
\end{tabular}}
\label{tab-task}
\end{table}

%% file: table/tab-exp-closed.tex
\begin{table}[t]
\setlength{\tabcolsep}{1.3pt}
\caption{Comparative evaluation of LLMs and MLLMs on \textit{DanmakuTPP-QA} closed-ended tasks. The table presents performance metrics (ACC: accuracy, higher is better; RMSE: root mean square error, lower is better) across 8 different tasks (T-1 through T-8) of traditional LLMs, MLLMs, and finetuned models, with specialized models achieving superior performance on specific tasks.}
\label{tab-exp-closed}
\resizebox{\linewidth}{!}{
\begin{tabular}{ccccccccc}
\toprule
\multicolumn{1}{c}{\textbf{Task}}                                                             & \textbf{T-1}         & \textbf{T-2}         & \textbf{T-3}         & \textbf{T-4}         & \textbf{T-5}         & \textbf{T-6}         & \textbf{T-7}         & \textbf{T-8}         \\ \hline
\multicolumn{1}{c}{\textbf{Model / Metrics}}                                                  & ACC~$\uparrow$                  & RMSE~$\downarrow$                 & RMSE~$\downarrow$                 & RMSE~$\downarrow$                 & RMSE~$\downarrow$                 & RMSE~$\downarrow$                 & ACC~$\uparrow$                  & ACC~$\uparrow$                  \\ \hline
\multicolumn{1}{c}{\textbf{LLMs}}                                                             & \multicolumn{1}{l}{} & \multicolumn{1}{l}{} & \multicolumn{1}{l}{} & \multicolumn{1}{l}{} & \multicolumn{1}{l}{} & \multicolumn{1}{l}{} & \multicolumn{1}{l}{} & \multicolumn{1}{l}{} \\
Qwen2.5-7B-Instruct~                                                                           & 0.33                 & 27.64               & 134.45              & 0.65                & 0.56                & 0.51                & 10.67                & 32.67                \\
Qwen2.5-32B-Instruct~                                                                          & 0.67                & 1.52                & 122.69              & 0.36                & 0.29                & 0.24                & 16.67                & 38.17                \\
Qwen2.5-72B-Instruct~                                                                          & 0.67                 & 1.28                & 123.45              & 0.30                & 0.46                & 0.46                & 16.00                 & 43.83                \\
Qwen3-8B~                                                                                      & 6.67                 & 1.80                & 123.59              & 0.32                & 0.41                & 0.45                & 19.33                & 41.50                 \\
Qwen3-30B-A3B~                                                                                  & 0.67                 & 1.33                & 121.96              & \textbf{0.20}                & 0.33                & 0.40                & \textbf{23.00}                 & 43.67                \\
Qwen3-235B-A22B~                                                                                & 8.67                 & 1.39                & \textbf{120.79}              & 0.30                & 0.31                & 0.29                & 10.33                & 32.50                 \\
Llama-3.3-70B-Instruct~                                                                       & 1.67                 & \textbf{1.11}                & 121.49              & 0.26                & 0.27                & 0.22                & 17.00                 & 33.33                \\
DeepSeek-V3~                                                                                   & \textbf{25.00}                 & 1.30                & 121.30              & 0.34                & \textbf{0.26}                & 0.22                & 13.67                & 34.5                 \\ \hline
\multicolumn{1}{c}{\textbf{MLLMs}}                                                            &                      &                      &                      &                      &                      &                      &                      &                      \\
Qwen2.5-VL-7B~                                                                                 & 9.67                 & 11.61               & 124.99              & 0.46                & 0.82                & 0.66                & 8.33                 & 22.17                \\
Qwen2.5-VL-32B~                                                                                & 8.0                  & 1.26                & 124.02              & 0.35                & 0.51                & 0.38                & 12.67                & 22.17                \\
Qwen2.5-VL-72B~                                                                                & 0.33                 & 1.14                & 121.25              & 0.28                & 0.47                & 0.41                & 15.98                & \textbf{47.17}                \\
Gemma3-27B~                                                                                    & 0.33                 & 1.33                & 121.32              & 0.28                & 0.27                & \textbf{0.20}                & 15.67                & 36.17                \\
\midrule
\multicolumn{1}{c}{\textbf{Finetuned}}                                                            &                      &                      &                      &                      &                      &                      &                      &                      \\
\rowcolor[gray]{0.8}\begin{tabular}[c]{@{}l@{}}
Qwen2.5-VL-3B~\end{tabular} & 27.0                 & 1.35                 & 220.43              & 0.05                & 0.16                & 0.08                & 15.33                & 43.00                 \\
\bottomrule
\end{tabular}}
\end{table}

%% file: checklist.tex
\section*{NeurIPS Paper Checklist}

\begin{enumerate}

\item {\bf Claims}
    \item[] Question: Do the main claims made in the abstract and introduction accurately reflect the paper's contributions and scope?
    \item[] Answer: \answerYes{}
    \item[] Justification: {The main claims made in the abstract and introduction accurately reflect the paper's contributions and scope. See the Abstract and Introduction sections.}
    \item[] Guidelines:
    \begin{itemize}
        \item The answer NA means that the abstract and introduction do not include the claims made in the paper.
        \item The abstract and/or introduction should clearly state the claims made, including the contributions made in the paper and important assumptions and limitations. A No or NA answer to this question will not be perceived well by the reviewers. 
        \item The claims made should match theoretical and experimental results, and reflect how much the results can be expected to generalize to other settings. 
        \item It is fine to include aspirational goals as motivation as long as it is clear that these goals are not attained by the paper. 
    \end{itemize}

\item {\bf Limitations}
    \item[] Question: Does the paper discuss the limitations of the work performed by the authors?
    \item[] Answer: \answerYes{}
    \item[] Justification: {The limitation of the work is discussed in the last section.}
    \item[] Guidelines:
    \begin{itemize}
        \item The answer NA means that the paper has no limitation while the answer No means that the paper has limitations, but those are not discussed in the paper. 
        \item The authors are encouraged to create a separate "Limitations" section in their paper.
        \item The paper should point out any strong assumptions and how robust the results are to violations of these assumptions (e.g., independence assumptions, noiseless settings, model well-specification, asymptotic approximations only holding locally). The authors should reflect on how these assumptions might be violated in practice and what the implications would be.
        \item The authors should reflect on the scope of the claims made, e.g., if the approach was only tested on a few datasets or with a few runs. In general, empirical results often depend on implicit assumptions, which should be articulated.
        \item The authors should reflect on the factors that influence the performance of the approach. For example, a facial recognition algorithm may perform poorly when image resolution is low or images are taken in low lighting. Or a speech-to-text system might not be used reliably to provide closed captions for online lectures because it fails to handle technical jargon.
        \item The authors should discuss the computational efficiency of the proposed algorithms and how they scale with dataset size.
        \item If applicable, the authors should discuss possible limitations of their approach to address problems of privacy and fairness.
        \item While the authors might fear that complete honesty about limitations might be used by reviewers as grounds for rejection, a worse outcome might be that reviewers discover limitations that aren't acknowledged in the paper. The authors should use their best judgment and recognize that individual actions in favor of transparency play an important role in developing norms that preserve the integrity of the community. Reviewers will be specifically instructed to not penalize honesty concerning limitations.
    \end{itemize}

\item {\bf Theory assumptions and proofs}
    \item[] Question: For each theoretical result, does the paper provide the full set of assumptions and a complete (and correct) proof?
    \item[] Answer: \answerYes{}
    \item[] Justification: {For all theoretical results, the paper provides the corresponding proofs in the appendix.} 
    \item[] Guidelines:
    \begin{itemize}
        \item The answer NA means that the paper does not include theoretical results. 
        \item All the theorems, formulas, and proofs in the paper should be numbered and cross-referenced.
        \item All assumptions should be clearly stated or referenced in the statement of any theorems.
        \item The proofs can either appear in the main paper or the supplemental material, but if they appear in the supplemental material, the authors are encouraged to provide a short proof sketch to provide intuition. 
        \item Inversely, any informal proof provided in the core of the paper should be complemented by formal proofs provided in appendix or supplemental material.
        \item Theorems and Lemmas that the proof relies upon should be properly referenced. 
    \end{itemize}

    \item {\bf Experimental result reproducibility}
    \item[] Question: Does the paper fully disclose all the information needed to reproduce the main experimental results of the paper to the extent that it affects the main claims and/or conclusions of the paper (regardless of whether the code and data are provided or not)?
    \item[] Answer: \answerYes{}
    \item[] Justification: {The paper fully discloses all the information needed to reproduce the main experimental results in the paper. See the Experiments section.}
    \item[] Guidelines:
    \begin{itemize}
        \item The answer NA means that the paper does not include experiments.
        \item If the paper includes experiments, a No answer to this question will not be perceived well by the reviewers: Making the paper reproducible is important, regardless of whether the code and data are provided or not.
        \item If the contribution is a dataset and/or model, the authors should describe the steps taken to make their results reproducible or verifiable. 
        \item Depending on the contribution, reproducibility can be accomplished in various ways. For example, if the contribution is a novel architecture, describing the architecture fully might suffice, or if the contribution is a specific model and empirical evaluation, it may be necessary to either make it possible for others to replicate the model with the same dataset, or provide access to the model. In general. releasing code and data is often one good way to accomplish this, but reproducibility can also be provided via detailed instructions for how to replicate the results, access to a hosted model (e.g., in the case of a large language model), releasing of a model checkpoint, or other means that are appropriate to the research performed.
        \item While NeurIPS does not require releasing code, the conference does require all submissions to provide some reasonable avenue for reproducibility, which may depend on the nature of the contribution. For example
        \begin{enumerate}
            \item If the contribution is primarily a new algorithm, the paper should make it clear how to reproduce that algorithm.
            \item If the contribution is primarily a new model architecture, the paper should describe the architecture clearly and fully.
            \item If the contribution is a new model (e.g., a large language model), then there should either be a way to access this model for reproducing the results or a way to reproduce the model (e.g., with an open-source dataset or instructions for how to construct the dataset).
            \item We recognize that reproducibility may be tricky in some cases, in which case authors are welcome to describe the particular way they provide for reproducibility. In the case of closed-source models, it may be that access to the model is limited in some way (e.g., to registered users), but it should be possible for other researchers to have some path to reproducing or verifying the results.
        \end{enumerate}
    \end{itemize}

\item {\bf Open access to data and code}
    \item[] Question: Does the paper provide open access to the data and code, with sufficient instructions to faithfully reproduce the main experimental results, as described in supplemental material?
    \item[] Answer: \answerYes{}
    \item[] Justification: {We provide the data and code in the supplemental material to reproduce the main experimental results.} 
    \item[] Guidelines:
    \begin{itemize}
        \item The answer NA means that paper does not include experiments requiring code.
        \item Please see the NeurIPS code and data submission guidelines (\url{https://nips.cc/public/guides/CodeSubmissionPolicy}) for more details.
        \item While we encourage the release of code and data, we understand that this might not be possible, so “No” is an acceptable answer. Papers cannot be rejected simply for not including code, unless this is central to the contribution (e.g., for a new open-source benchmark).
        \item The instructions should contain the exact command and environment needed to run to reproduce the results. See the NeurIPS code and data submission guidelines (\url{https://nips.cc/public/guides/CodeSubmissionPolicy}) for more details.
        \item The authors should provide instructions on data access and preparation, including how to access the raw data, preprocessed data, intermediate data, and generated data, etc.
        \item The authors should provide scripts to reproduce all experimental results for the new proposed method and baselines. If only a subset of experiments are reproducible, they should state which ones are omitted from the script and why.
        \item At submission time, to preserve anonymity, the authors should release anonymized versions (if applicable).
        \item Providing as much information as possible in supplemental material (appended to the paper) is recommended, but including URLs to data and code is permitted.
    \end{itemize}

\item {\bf Experimental setting/details}
    \item[] Question: Does the paper specify all the training and test details (e.g., data splits, hyperparameters, how they were chosen, type of optimizer, etc.) necessary to understand the results?
    \item[] Answer: \answerYes{}
    \item[] Justification: {We specify all the training and test details necessary to understand the results. See the Experiments section.} 
    \item[] Guidelines:
    \begin{itemize}
        \item The answer NA means that the paper does not include experiments.
        \item The experimental setting should be presented in the core of the paper to a level of detail that is necessary to appreciate the results and make sense of them.
        \item The full details can be provided either with the code, in appendix, or as supplemental material.
    \end{itemize}

\item {\bf Experiment statistical significance}
    \item[] Question: Does the paper report error bars suitably and correctly defined or other appropriate information about the statistical significance of the experiments?
    \item[] Answer: \answerYes{}
    \item[] Justification: {We report the statistical significance of the experiments in Figure 2.} 
    \item[] Guidelines:
    \begin{itemize}
        \item The answer NA means that the paper does not include experiments.
        \item The authors should answer "Yes" if the results are accompanied by error bars, confidence intervals, or statistical significance tests, at least for the experiments that support the main claims of the paper.
        \item The factors of variability that the error bars are capturing should be clearly stated (for example, train/test split, initialization, random drawing of some parameter, or overall run with given experimental conditions).
        \item The method for calculating the error bars should be explained (closed form formula, call to a library function, bootstrap, etc.)
        \item The assumptions made should be given (e.g., Normally distributed errors).
        \item It should be clear whether the error bar is the standard deviation or the standard error of the mean.
        \item It is OK to report 1-sigma error bars, but one should state it. The authors should preferably report a 2-sigma error bar than state that they have a 96\% CI, if the hypothesis of Normality of errors is not verified.
        \item For asymmetric distributions, the authors should be careful not to show in tables or figures symmetric error bars that would yield results that are out of range (e.g. negative error rates).
        \item If error bars are reported in tables or plots, The authors should explain in the text how they were calculated and reference the corresponding figures or tables in the text.
    \end{itemize}

\item {\bf Experiments compute resources}
    \item[] Question: For each experiment, does the paper provide sufficient information on the computer resources (type of compute workers, memory, time of execution) needed to reproduce the experiments?
    \item[] Answer: \answerYes{}
    \item[] Justification: {All experiments are conducted on the server whose details are provided in the paper. See the Experiments section.}
    \item[] Guidelines:
    \begin{itemize}
        \item The answer NA means that the paper does not include experiments.
        \item The paper should indicate the type of compute workers CPU or GPU, internal cluster, or cloud provider, including relevant memory and storage.
        \item The paper should provide the amount of compute required for each of the individual experimental runs as well as estimate the total compute. 
        \item The paper should disclose whether the full research project required more compute than the experiments reported in the paper (e.g., preliminary or failed experiments that didn't make it into the paper). 
    \end{itemize}
    
\item {\bf Code of ethics}
    \item[] Question: Does the research conducted in the paper conform, in every respect, with the NeurIPS Code of Ethics \url{https://neurips.cc/public/EthicsGuidelines}?
    \item[] Answer: \answerYes{}
    \item[] Justification: {The research conducted in the paper conform with the NeurIPS Code of Ethics.}
    \item[] Guidelines:
    \begin{itemize}
        \item The answer NA means that the authors have not reviewed the NeurIPS Code of Ethics.
        \item If the authors answer No, they should explain the special circumstances that require a deviation from the Code of Ethics.
        \item The authors should make sure to preserve anonymity (e.g., if there is a special consideration due to laws or regulations in their jurisdiction).
    \end{itemize}

\item {\bf Broader impacts}
    \item[] Question: Does the paper discuss both potential positive societal impacts and negative societal impacts of the work performed?
    \item[] Answer: \answerNo{} 
    \item[] Justification: This paper presents work whose goal is to advance the field of machine learning. There are many potential societal consequences of our work, none of which we feel must be specifically highlighted here. 
    \item[] Guidelines:
    \begin{itemize}
        \item The answer NA means that there is no societal impact of the work performed.
        \item If the authors answer NA or No, they should explain why their work has no societal impact or why the paper does not address societal impact.
        \item Examples of negative societal impacts include potential malicious or unintended uses (e.g., disinformation, generating fake profiles, surveillance), fairness considerations (e.g., deployment of technologies that could make decisions that unfairly impact specific groups), privacy considerations, and security considerations.
        \item The conference expects that many papers will be foundational research and not tied to particular applications, let alone deployments. However, if there is a direct path to any negative applications, the authors should point it out. For example, it is legitimate to point out that an improvement in the quality of generative models could be used to generate deepfakes for disinformation. On the other hand, it is not needed to point out that a generic algorithm for optimizing neural networks could enable people to train models that generate Deepfakes faster.
        \item The authors should consider possible harms that could arise when the technology is being used as intended and functioning correctly, harms that could arise when the technology is being used as intended but gives incorrect results, and harms following from (intentional or unintentional) misuse of the technology.
        \item If there are negative societal impacts, the authors could also discuss possible mitigation strategies (e.g., gated release of models, providing defenses in addition to attacks, mechanisms for monitoring misuse, mechanisms to monitor how a system learns from feedback over time, improving the efficiency and accessibility of ML).
    \end{itemize}
    
\item {\bf Safeguards}
    \item[] Question: Does the paper describe safeguards that have been put in place for responsible release of data or models that have a high risk for misuse (e.g., pretrained language models, image generators, or scraped datasets)?
    \item[] Answer: \answerNA{}
    \item[] Justification: {The paper poses no such risks.}
    \item[] Guidelines:
    \begin{itemize}
        \item The answer NA means that the paper poses no such risks.
        \item Released models that have a high risk for misuse or dual-use should be released with necessary safeguards to allow for controlled use of the model, for example by requiring that users adhere to usage guidelines or restrictions to access the model or implementing safety filters. 
        \item Datasets that have been scraped from the Internet could pose safety risks. The authors should describe how they avoided releasing unsafe images.
        \item We recognize that providing effective safeguards is challenging, and many papers do not require this, but we encourage authors to take this into account and make a best faith effort.
    \end{itemize}

\item {\bf Licenses for existing assets}
    \item[] Question: Are the creators or original owners of assets (e.g., code, data, models), used in the paper, properly credited and are the license and terms of use explicitly mentioned and properly respected?
    \item[] Answer: \answerYes{}
    \item[] Justification: {All code, models, and datasets mentioned in the text are appropriately cited with their original papers.} 
    \item[] Guidelines:
    \begin{itemize}
        \item The answer NA means that the paper does not use existing assets.
        \item The authors should cite the original paper that produced the code package or dataset.
        \item The authors should state which version of the asset is used and, if possible, include a URL.
        \item The name of the license (e.g., CC-BY 4.0) should be included for each asset.
        \item For scraped data from a particular source (e.g., website), the copyright and terms of service of that source should be provided.
        \item If assets are released, the license, copyright information, and terms of use in the package should be provided. For popular datasets, \url{paperswithcode.com/datasets} has curated licenses for some datasets. Their licensing guide can help determine the license of a dataset.
        \item For existing datasets that are re-packaged, both the original license and the license of the derived asset (if it has changed) should be provided.
        \item If this information is not available online, the authors are encouraged to reach out to the asset's creators.
    \end{itemize}

\item {\bf New assets}
    \item[] Question: Are new assets introduced in the paper well documented and is the documentation provided alongside the assets?
    \item[] Answer: \answerYes{}
    \item[] Justification: {New assets introduced in the paper, in particular the code and datasets, are well documented. The documentation is provided through online platforms including Huggingface and Github.}
    \item[] Guidelines:
    \begin{itemize}
        \item The answer NA means that the paper does not release new assets.
        \item Researchers should communicate the details of the dataset/code/model as part of their submissions via structured templates. This includes details about training, license, limitations, etc. 
        \item The paper should discuss whether and how consent was obtained from people whose asset is used.
        \item At submission time, remember to anonymize your assets (if applicable). You can either create an anonymized URL or include an anonymized zip file.
    \end{itemize}

\item {\bf Crowdsourcing and research with human subjects}
    \item[] Question: For crowdsourcing experiments and research with human subjects, does the paper include the full text of instructions given to participants and screenshots, if applicable, as well as details about compensation (if any)? 
    \item[] Answer: \answerNA{}
    \item[] Justification: {The paper does not involve crowdsourcing nor research with human subjects.}
    \item[] Guidelines:
    \begin{itemize}
        \item The answer NA means that the paper does not involve crowdsourcing nor research with human subjects.
        \item Including this information in the supplemental material is fine, but if the main contribution of the paper involves human subjects, then as much detail as possible should be included in the main paper. 
        \item According to the NeurIPS Code of Ethics, workers involved in data collection, curation, or other labor should be paid at least the minimum wage in the country of the data collector. 
    \end{itemize}

\item {\bf Institutional review board (IRB) approvals or equivalent for research with human subjects}
    \item[] Question: Does the paper describe potential risks incurred by study participants, whether such risks were disclosed to the subjects, and whether Institutional Review Board (IRB) approvals (or an equivalent approval/review based on the requirements of your country or institution) were obtained?
    \item[] Answer: \answerNA{}
    \item[] Justification: {The paper does not involve crowdsourcing nor research with human subjects.}
    \item[] Guidelines:
    \begin{itemize}
        \item The answer NA means that the paper does not involve crowdsourcing nor research with human subjects.
        \item Depending on the country in which research is conducted, IRB approval (or equivalent) may be required for any human subjects research. If you obtained IRB approval, you should clearly state this in the paper. 
        \item We recognize that the procedures for this may vary significantly between institutions and locations, and we expect authors to adhere to the NeurIPS Code of Ethics and the guidelines for their institution. 
        \item For initial submissions, do not include any information that would break anonymity (if applicable), such as the institution conducting the review.
    \end{itemize}

\item {\bf Declaration of LLM usage}
    \item[] Question: Does the paper describe the usage of LLMs if it is an important, original, or non-standard component of the core methods in this research? Note that if the LLM is used only for writing, editing, or formatting purposes and does not impact the core methodology, scientific rigorousness, or originality of the research, declaration is not required.
    \item[] Answer: \answerNA{}
    \item[] Justification: {The core method development in this research does not involve LLMs as any important, original, or non-standard components.}
    \item[] Guidelines:
    \begin{itemize}
        \item The answer NA means that the core method development in this research does not involve LLMs as any important, original, or non-standard components.
        \item Please refer to our LLM policy (\url{https://neurips.cc/Conferences/2025/LLM}) for what should or should not be described.
    \end{itemize}

\end{enumerate}

%% file: Appendix.tex
\section{Ethical Compliance Measures}
\label{appendix-ethical}
During the data collection process, our handling of copyright, privacy, and harmful content aligns with academic norms and legal frameworks:

\textbf{Copyright and Data Usage Compliance.} All video data we used is collected from the Bilibili platform, which allows public access to videos and Danmaku (bullet comments) via its front-end interfaces. To ensure compliance: (1) we only collected data from publicly available videos without any login or authentication; (2) our dataset does not contain any raw video data or audio. Instead, we only extract frame-level visual features and metadata of Danmaku comments, including timestamp and text. Full videos are neither stored nor reconstructible; (3) the dataset is released strictly for academic and non-commercial research, strictly adhering to the Bilibili Platform Terms of Service.

\textbf{Data Privacy and User Anonymization.}
We take user privacy seriously: (1) No user IDs, usernames, IP addresses, or personal identifiers are collected or released. (2) Danmaku comments are publicly visible and are usually pseudonymous or anonymous. All Danmaku comments within the interactive videos are devoid of any association with the identity information of the users who posted them. The dataset we constructed only contains Danmaku text, temporal sequences, and video frame images, and does not include any user account information, personally identifiable information (PII), or other privacy-sensitive content associated with users. (3) We perform automatic filtering to remove any Danmaku containing potential personal information or sensitive content (e.g., email addresses, phone numbers). Additionally, as demonstrated in our multi-agent collaborative framework, we have implemented a manual review phase after data construction was completed to filter out potentially policy-related data. (4) We do not perform user profiling or behavioral tracking; our focus is solely on timestamped event-level modeling.

\textbf{Harassment or Harmful Content.}
All interactive videos and the accompanying Danmaku text underwent rigorous content moderation upon upload to the Bilibili platform, strictly adhering to its user agreements and relevant regional laws and regulations.
In addition to relying on Bilibili’s Terms of Service and built-in moderation systems, we actively implemented multiple methodological precautions to ensure ethical compliance. During the data collection and sanitization stage:
(1) We construct an open-source stopword list, which includes terms related to advertisements, curse, drug, gamble, polity, porn, violence, sensitive/phishing URLs, and other sensitive categories, to filter out Danmaku events containing inappropriate content. The sensitive/dirty stopword list we employed (primarily Chinese, with some English expressions included) was constructed by merging several open-source sensitive/dirty stopword lists from GitHub, which ensures the comprehensiveness of our stopword list.
(2) Rather than randomly sampling videos from the platform, we deliberately curated our dataset from high-quality sources. Specifically, we selected all publicly available, open-access videos released by the Top-100 most influential content creators on the platform, as recognized by the platform and affiliated media outlets in the corresponding year. These videos—and their associated Danmaku comments—have been subject to sustained audience scrutiny and extensive platform moderation over time, making them among the cleanest and most representative content on the platform. This selective data sourcing strategy laid the foundation for building a dataset with minimal ethical concerns.
(3) Since DanmakuTPP-QA was constructed using a multi-agent framework, we recognize the potential risk of the model introducing bias or generating harmful content. To mitigate this, we engaged three graduate students to manually review the entirety of the DanmakuTPP-QA dataset.

\section{Details of Multi-agent Framework}
\label{appendix-a}
\begin{figure}[t]
    \centering
    \includegraphics[width=\textwidth]{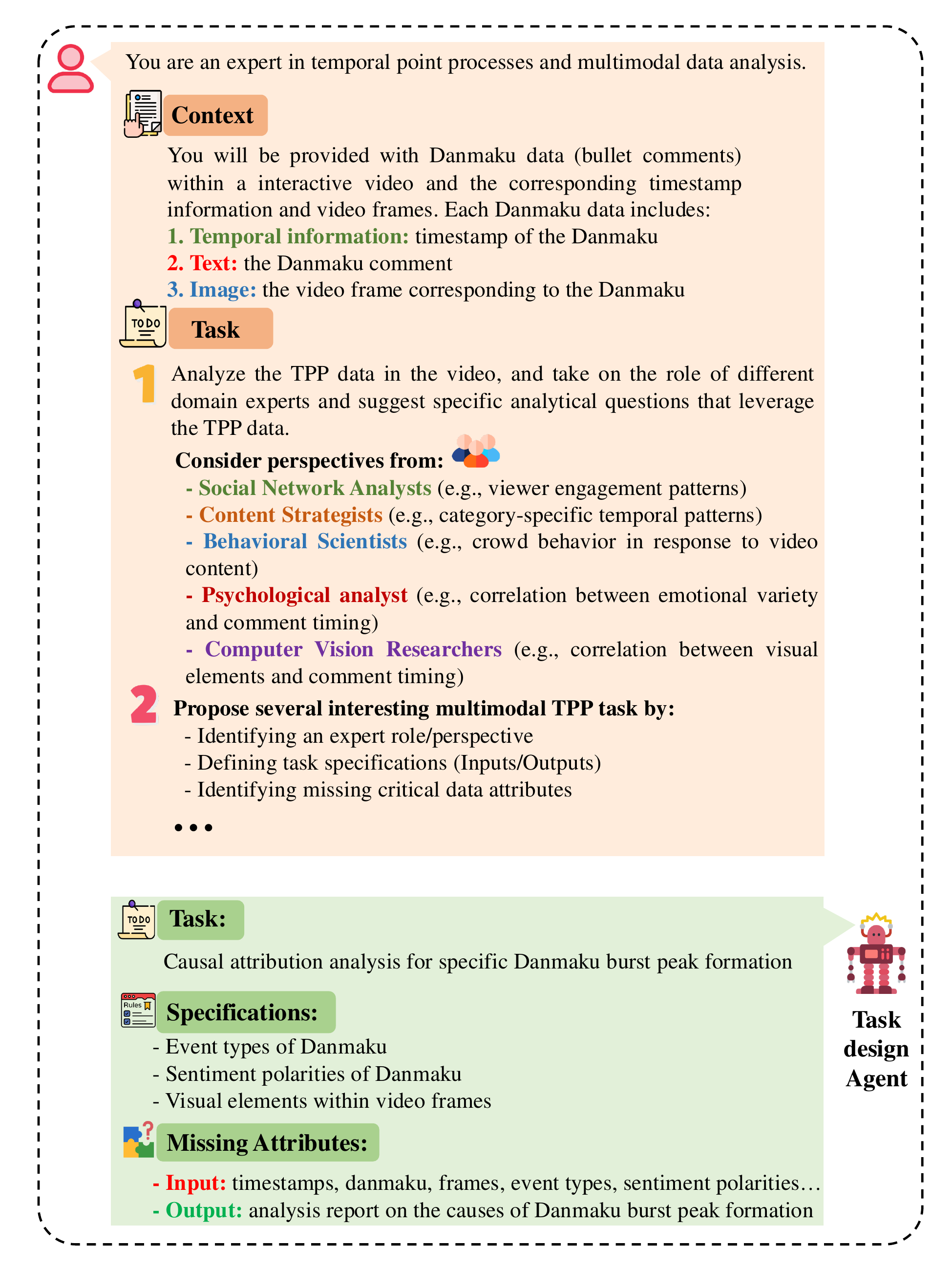}
    \caption{Prompt template for Task-design Agent and its corresponding outputs.}
    \label{fig-task-design}
\end{figure}

\textbf{Task-design Agent.} 
We employ DeepSeek-R1 as the Task-design Agent to simulate experts from diverse domains, aiming to explore potential meaningful research tasks of danmaku temporal point process data. Additionally, the module is also responsible for defining formal specifications for designed tasks and analyzing the deficiencies in current datasets regarding required attributes for task resolution. The prompt template guiding the Task-design Agent's operations, along with a representative output sample, is systematically presented in Figure~\ref{fig-task-design}.

\begin{figure}[t]
    \centering
    \includegraphics[width=\textwidth]{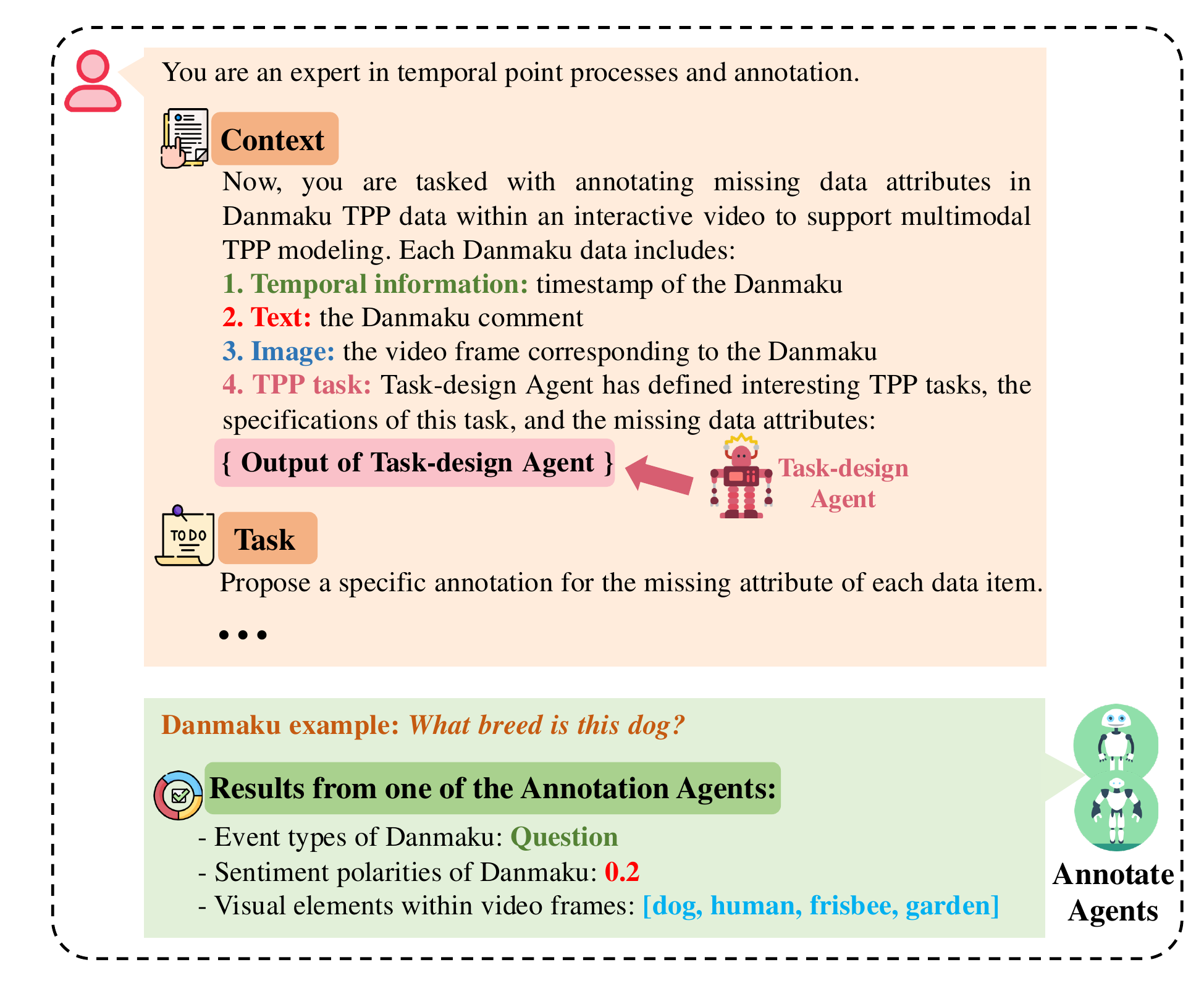}
    \caption{Prompt template for Annotation Agents and corresponding annotation results.}
    \label{fig-annotation}
\end{figure}

\textbf{Annotation Agent.}
As described in the main text, this module serves to identify deficient attributes within the dataset, thereby supporting the resolution of research tasks designed in the preceding module. Within the Annotation Agent group, Qwen2.5 handles textual annotation while Qwen2.5-VL and RAM manage visual annotation. The guiding prompt template and representative annotation results are illustrated in Figure~\ref{fig-annotation}.

\begin{figure}[t]
    \centering
    \includegraphics[width=\textwidth]{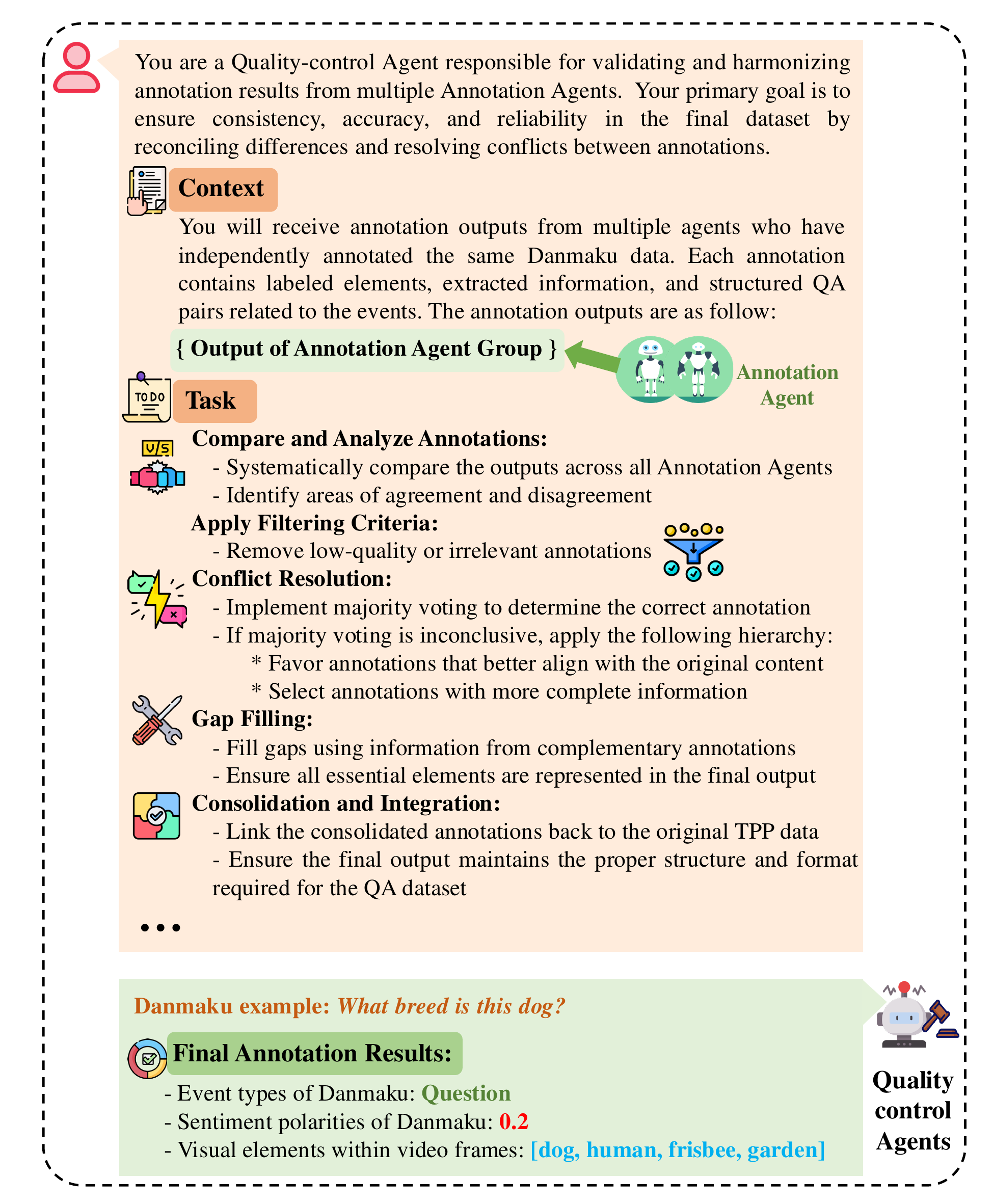}
    \caption{Prompt template for Quality-control Agent and its corresponding output.}
    \label{fig-quality-control}
\end{figure}

\textbf{Quality-control Agent.}
We utilize Qwen3 as the Quality-control Agent to aggregate and reconcile the outputs from multiple Annotation Agents, effectively resolving discrepancies arising from conflicting annotations. The prompt for this module and the output samples are illustrated in Figure~\ref{fig-quality-control}.

\begin{figure}[t]
    \centering
    \includegraphics[width=\textwidth]{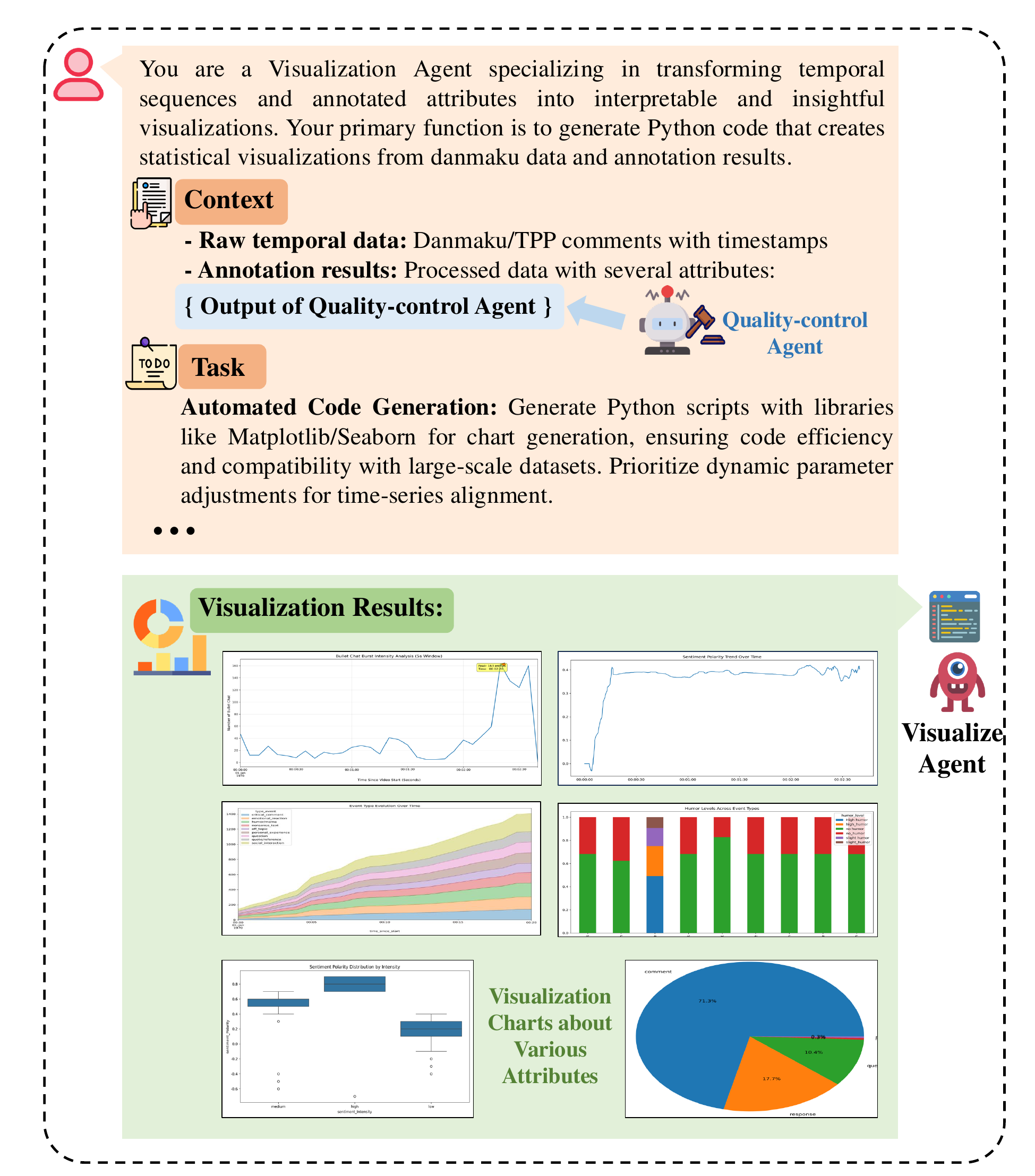}
    \caption{Prompt template for Visualization Agent and various corresponding charts.}
    \label{fig-visualization}
\end{figure}

\textbf{Visualization Agent.}
This module is responsible for transforming temporal sequences and annotated attributes into interpretable visualizations. The prompt template used to guide the agent and the visualization chart of the outputs are shown in Figure~\ref{fig-visualization}.

\begin{figure}[t]
    \centering
    \includegraphics[width=\textwidth]{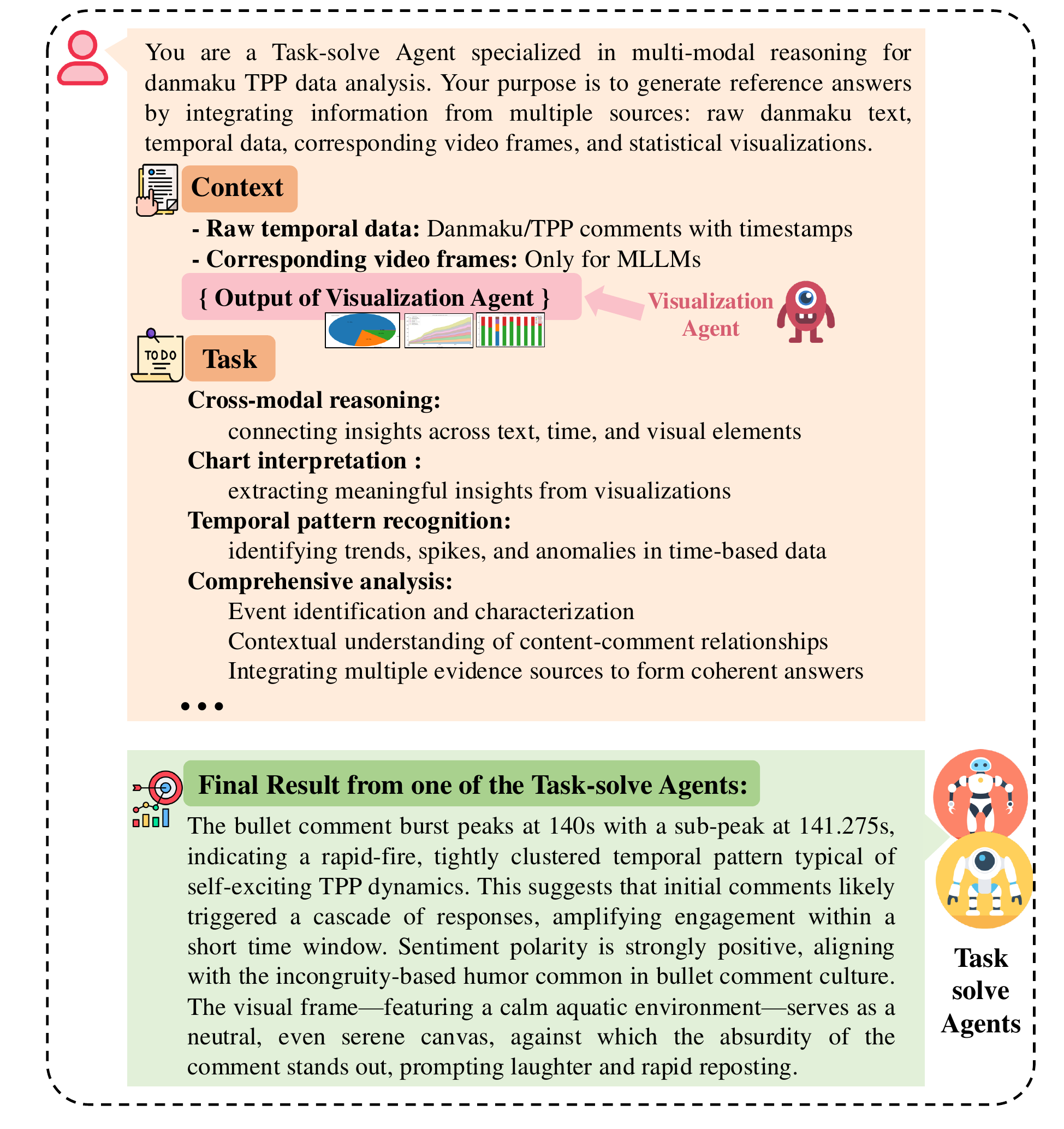}
    \caption{Prompt template for Task-solve Agents and the corresponding output.}
    \label{fig-task-solve}
\end{figure}

\textbf{Task-solve Agent.} 
We incorporate multiple robust LLMs and MLLMs in the Task-solve Agent group to resolve the tasks and generate reference answers as ground-truth labels in DanmakuTPP-QA. The prompt template of this module and the output samples are depicted in Figure~\ref{fig-task-solve}.
By orchestrating the above agents in a coordinated sequence, the multi-agent system enables the systematic construction of high-quality datasets specifically engineered for multimodal temporal modeling and TPP understanding.

\section{Details of Finetuning Experiment}
\label{appendix-b}
As described in the main manuscript, we evaluate the performance of the Qwen2.5-VL-3B model after LoRA finetuning on the DanmakuTPP-QA dataset. For each task, training is conducted on a single NVIDIA RTX 4090 GPU for 3 epochs using a learning rate of 1e-4, a batch size of 1, and gradient accumulation steps of 4. The LoRA configuration employed a rank of 64, an alpha value of 16, and a dropout rate of 0.05. Due to GPU memory constraints, we employ left sequence truncation during training, which preserves the more recent right-side context containing the most task-relevant information. The maximum sequence length after truncation varies by task and is specified in Table~\ref{tab-appendix}.
\input{table/tab-appendix}

\newpage

%% file: table/tab-appendix.tex
\begin{table}[h]
\caption{The task-specific maximum sequence lengths after left truncation during finetuning.}
\centering
\resizebox{\textwidth}{!}{
    \begin{tabular}{ccccccccccc}
        \hline
        \textbf{Task}                & \textbf{T-1} & \textbf{T-2} & \textbf{T-3} & \textbf{T-4} & \textbf{T-5} & \textbf{T-6} & \textbf{T-7} & \textbf{T-8} & \textbf{T-9} & \textbf{T-10} \\ \hline
        max sequence length & 100 & 200 & 200 & 50  & 60  & 60  & 50  & 50  & 100 & 100  \\ \hline
    \end{tabular}
}
\label{tab-appendix}
\end{table}